\documentclass{article}
\pdfoutput=1

\usepackage[preprint, nonatbib]{neurips_2024}

\usepackage[utf8]{inputenc} % allow utf-8 input
\usepackage[T1]{fontenc}    % use 8-bit T1 fonts
\usepackage{hyperref}       % hyperlinks
\usepackage{url}            % simple URL typesetting
\usepackage{booktabs}       % professional-quality tables
\usepackage{amsfonts}       % blackboard math symbols
\usepackage{nicefrac}       % compact symbols for 1/2, etc.
\usepackage{microtype}      % microtypography
\usepackage{xcolor}         % colors
\usepackage{graphicx} 
\usepackage{multicol}
\usepackage{multirow}
\usepackage[numbers]{natbib}

\usepackage{amsmath,amssymb,amsfonts}
\usepackage[ruled]{algorithm2e}
\usepackage{amsthm}
\newtheorem{assumption}{Assumption}
\newtheorem{theorem}{Theorem}
\newtheorem{definition}{Definition}
\newtheorem{corollary}{Corollary}

\title{Variance Control for Black Box Variational Inference Using The James-Stein Estimator}

\author{%
  Dominic B.~Dayta \\
  Mathematical Informatics\\
  Nara Institute of Science and Technology\\
  \texttt{dominic.dayta.da4@naist.ac.jp} \\
}

\begin{document}

\maketitle

\begin{abstract}
Black Box Variational Inference is a promising framework in a succession of recent efforts to make Variational Inference more ``black box". However, in basic version it either fails to converge due to instability or requires some fine-tuning of the update steps prior to execution that hinder it from being completely general purpose. We propose a method for regulating its parameter updates by reframing stochastic gradient ascent as a multivariate estimation problem. We examine the properties of the James-Stein estimator as a replacement for the arithmetic mean of Monte Carlo estimates of the gradient of the evidence lower bound. The proposed method provides relatively weaker variance reduction than Rao-Blackwellization, but offers a tradeoff of being simpler and requiring no fine tuning on the part of the analyst. Performance on benchmark datasets also demonstrate a consistent performance at par or better than the Rao-Blackwellized approach in terms of model fit and time to convergence.
\end{abstract}

\section{Introduction}

Black Box Variational Inference (BBVI) \cite{Ranganath-2014} presents a promising alternative to MCMC-based techniques for fitting the posterior distribution in arbitrarily large Bayesian models. In line with the general framework of Variational Inference (VI) \cite{Blei-2017}, BBVI works around the tendency of MCMC solutions to explode in computational complexity by providing an approximate, instead of an exact, solution to the optimal parameters defining the model.

Where exact MCMC methods aim to produce random samples of the posterior distribution, VI works by approximating the posterior with a family of tractable densities, indexed by free parameters known as \textit{variational parameters}. In this way, VI changes the problem from one of sampling (as in MCMC) to one of optimization, to find the variational parameters that make the resulting density as close as possible (in terms of Kullback-Leibler divergence) to the target posterior distribution \cite{Blei-2017}.

However, the process of finding the correct optimization algorithm for the variational parameters can lead to highly complex derivations that are prone to human error, still rendering the process of model exploration quite slow \cite{Blei-2017}. The promise of BBVI is in removing the need for such derivations, by providing a generalized algorithm for finding the variational parameters for models of any form and size.

Our contribution lies primarily in our demonstration of the variance reduction properties of using the James-Stein estimator in estimating noisy ELBO gradients in the BBVI Update Steps. Through simulation studies, we are able to conclude that the James-Stein estimator yields tightly controlled variance of the noisy ELBO gradients even as the complexity of the model grows.

The rest of this paper is organized as follows: section \ref{sec:bbvi} presents the problem setting, introducing the general algorithm for BBVI as well as its improved version via Rao-Blackwellization. We then present in section \ref{sec:stein} our proposed improvement using the James-Stein estimator, casting BBVI (and in general, stochastic gradient ascent problems) in the language of classical point estimation theory. Finally, we provide empirical results in \ref{sec:gausmix} through simulated and benchmark datasets involving finite mixtures of Gaussians.

\section{Preliminaries}
\label{sec:bbvi}

We briefly introduce the general algorithm for BBVI. Suppose we have data $y = \{y_1, y_2,..., y_n\}$ with $n$ observations for which we have posited some arbitrary model parameterized by $\theta$, with prior density $p(\theta)$. Bayesian data analysis primarily makes use of the posterior density,
$$
p(\theta | y) = p(y | \theta) p(\theta) / p(y)
$$
where $p(y | \theta)$ denotes the model likelihood, and the marginal distribution $p(y)$ is known as the "evidence''.

For models arbitrarily large (or in the case of non-conjugacy), finding the form of $p(\theta | y)$ can be exceedingly complex. In many cases, $p(\theta | y)$ may not even belong to a known family of densities, and the complexity of the model makes applying exact MCMC inference too slow, consequently inhibiting frequent and comprehensive model criticism and exploration. VI proceeds by finding an approximating distribution $q(\theta | \lambda)$ that is much simpler in form, but analytically is nearly identical to the posterior distribution. This is achieved by allowing the parameter $\lambda$ to vary freely, and finding its value such that the Kullback-Leibler divergence,
\begin{align}
\label{eq:kullback}
KL(q(\theta | \lambda)||p(\theta|y)) = \int_{\theta} \log\bigg( \frac{q(\theta | \lambda)}{p(\theta|y)} q(\theta | \lambda) \bigg)d\theta
\end{align}
is minimized.

Optimizing the KL-divergence directly can be intractable. Alternatively, one can make use of a quantity referred to in VI and Expectation Propagation literature as the \textit{evidence lower-bound} (ELBO), maximizing on which is equivalent to minimizing the KL-divergence \cite{Ranganath-2014}. Via a common decomposition \cite{Goodfellow-2016} of (\ref{eq:kullback}), we can write the logarithm of the evidence $p(y)$ as
$$
\log p(y) = \mathcal{L}(\lambda) + KL(q(\theta | \lambda)||p(\theta|y))
$$
where
\begin{align}
\label{eq:elbo}
\mathcal{L}(\lambda) = \mathbb{E}_{q}[\log p(y,\theta) - \log q(\theta | \lambda) ]
\end{align}

The form (\ref{eq:elbo}) is called the evidence lower-bound as $\log p(y) = \mathcal{L}(\lambda)$ when $q$ equals the posterior distribution exactly (i.e., $KL(q(\theta | \lambda)||p(\theta|y)) = 0$), and $\log p(y) > \mathcal{L}(\lambda)$ otherwise. Thus, maximizing $\mathcal{L}(\lambda)$ is equivalent to minimizing on $KL(q(\theta | \lambda)||p(\theta|y))$.

\subsection{Black Box Variational Inference}

Performing VI typically requires finding the appropriate coordinate ascent algorithm needed specific to each model combination \cite{Blei-2017}, but in BBVI this step is conveniently left out, opting instead of a general algorithm that works for most cases. It is shown \cite{Ranganath-2014} that the gradient of $\mathcal{L}(\lambda)$ is given by
\begin{align}
\label{eq:elbograd}
\nabla_\lambda \mathcal{L}(\lambda) = \mathbb{E}_{q}[\nabla_{\lambda} \log q(\theta | \lambda) (\log p(y, \theta) - \log q(\theta | \lambda))]
\end{align}

Which can now be used in a general gradient ascent algorithm. A sample of $S$ draws $\theta \sim q(\theta | \lambda)$ can be obtained and used to estimate the expectation using
\begin{align}
\label{eq:naivebbvi}
    \hat\nabla_\lambda \mathcal{L} = \frac{1}{S} \sum^{S}_{s=1} \nabla_{\lambda} \log q(\theta[s] | \lambda) (\log p(y, \theta[s]) - \log q(\theta[s] | \lambda))
\end{align}

In simpler cases the gradient $\nabla_{\lambda} \log q(\theta | \lambda)$, also known in classical statistical theory as the \textit{score function}, can be obtained analytically, but various autodifferentiation packages have become available for most computational environments such that the algorithm can truly be approached in a black-box manner. We can now present the ``Naive'' form of BBVI in terms of the stochastic gradient ascent formulation in Algorithm \ref{alg:bbvi} in our Supplementary Materials.

\subsection{Variance Control Using Rao-Blackwellization}

However, this form of BBVI is noted for either failing to converge or find meaningful solutions within reasonable time due to a high variance in its sampling distribution. This problem is addressed in the original paper through Rao-Blackwellization \cite{Casella-1996}. First, we note that in most applications there will be more than one parameter $\theta_1, \theta_2, ..., \theta_p$ in the posterior distribution, and generally each of them are assigned their corresponding variational distribution $\lambda_1, \lambda_2, ..., \lambda_p$. The simplest and most commonly used \cite{Blei-2017} variational family is the \textit{mean-field} family, defined as follows.

\begin{definition}
  $Q$ is known as the \emph{mean-field variational family} if for all $q \in Q$:
  \begin{align}
      q(\theta | \lambda) = \prod^{p}_{i=1} q(\theta_i | \lambda_i)
  \end{align}
\end{definition}

To be sure, the mean-field family is not the only one used when performing either BBVI or VI. Other forms of VI have been proposed and explored in the literature for specific cases where the mean-field assumption may be inappropriate \cite{Blei-2017}. For instance, \textit{structured variational inference} \cite{Saul-1995, Barber-1998} removes the independence assumption that is inherent in the mean-field flavor family by specifically inducing dependencies between the variational parameters. Another approach is to expand the mean-field family with the addition of latent variables that encode these relationships \cite{Bishop-1997}. We follow the mean-field assumption to facilitate a straightforward optimization problem in this paper, although our proposed black box approach does not specifically require this factorization to hold.

In performing Rao-Blackwellization this factorization is exploited through the average (\ref{eq:naivebbvi}), specifically the difference between the log-joint distribution $\log p(y, \theta[s])$ and its approximation $\log q(\theta[s] | \lambda)$, being equal to
\begin{align*}
    \log p(y, \theta[s]) - \log q(\theta[s] | \lambda) = \sum^{p}_{j=1} (\log p(y, \theta_{\lambda_j}[s]) - \log q(\theta_{\lambda_j}[s] | \lambda_j))
\end{align*}
This grows linearly with the number of variational parameters $p$. Hence, the variance of the gradient estimate $\hat{\nabla}_\lambda\mathcal{L}$ grows linearly as well. More importantly, we see that this growth in variance is mostly unnecessary, as updating a particular parameter $\lambda_j$ will be based on a gradient whose variance is composed of those for other parameters $\lambda_{j'}$, $j' \neq j$.

Using Rao-Blackwellization, the gradient is estimated for each $\lambda_j$ parameter conditioned on current values of the other variational parameters. That is, given $q(\theta_{\lambda_j} | \lambda_j)$ as the terms of the approximation that depend only on $\lambda_j$, and $p(y, \theta_{\lambda_j})$ as the terms of the joint distribution keeping only $\theta_{\lambda_j}$ depending on $\lambda_j$. Using an analogy from graphical models, these remaining terms are referred to as the \textit{Markov Blanket} \cite{Pearl-1988} of $\lambda_j$. The update rule is modified in Algorithm \ref{alg:bbvi-rb} in our Supplementary Materials.

This results in best-in-class variance reduction for the algorithm. However, factorizing the joint and variational distributions to obtain the corresponding update steps, while straightforward, requires additional steps for the analyst before conducting BBVI, and can become unnecessarily tedious in the case of large, highly-layered Bayesian models. This step is a significant hurdle in achieving a truly \textit{black-box} algorithm, and so in Section \ref{sec:stein}, we discuss an alternative that does not require finding the appropriate factorization. We first conclude our discussion of preliminaries with a brief overview of some related work and recent publications that have appeared since \cite{Ranganath-2014}.

\subsection{Related Work}

BBVI has received increasing attention in the machine learning literature for its promise of a general algorithm that can be applicable in a wide variety of settings. Recent work have explored improvements to the algorithm by providing adaptive stopping criteria \cite{Welandawe-2022}, as well as proving convergence guarantees within common expected scenarios \cite{Domke-2020}.

A related work worth mentioning \cite{Xu-2019} examines using the reparameterization trick \cite{Kingma-2013, Rezende-2014} to reduce the variance of the gradient estimates used for the algorithm's update step. However, the reparametrization trick is not strictly a method for variance control in BBVI. Instead, it is a method for changing the parameters of a learning parameter to remove constraints that might hinder computation. This means that both Rao-Blackwellization and the James-Stein estimator can be used on the gradient estimators under reparametrization. We therefore focus primarily on Rao-Blackwellization as a benchmark for our analysis.

Similarly, our proposed methodology remains sufficiently general such that it should be straightforward to combine with other emerging practices being proposed and tested in more recent works. For instance, it is possible to apply our regularized ELBO gradient estimate with the automated stopping and estimate correction logic proposed by \cite{Welandawe-2022}. In the following paper, we have returned to the basics of BBVI to ensure that any variance reduction observed can confidently be attributed to our proposed estimation method, and not as a side effect of competing layers in the algorithm.

Our proposal for the use of the James-Stein estimator is motivated by the logic of biasing the noisy gradient estimate in stochastic gradient descent/ascent problems by preventing exploding gradients from straying parameter updates into problematic regions. This is not a new idea in the deep learning literature, where heuristics like gradient clipping \cite{Koloskova-2023, Goodfellow-2016} are already established with proven convergence guarantees. We explore this connection further in section \ref{sec:gradclip}.

\section{Variance Control Using The James-Stein Estimator}
\label{sec:stein}

We now propose our James-Stein BBVI with the objective of performing general variational inference without requiring the analyst to find the necessary factorizations for BBVI-RB. This is achieved through the James-Stein estimator. To better understand the motivation behind this proposal, we recast the problem of gradient ascent in BBVI as an estimation problem.

\subsection{Gradient Ascent As Estimation}

In Algorithms \ref{alg:bbvi} and \ref{alg:bbvi-rb}, we perform gradient ascent of the form
\begin{align}
\label{eq:sgd}
    \lambda^t = \lambda^{t-1} + \rho^t \hat\nabla_\lambda\mathcal{L}(\lambda^{t-1})
\end{align}
where $\hat\nabla_\lambda\mathcal{L}(\lambda^{t-1})$ is obtained via Monte Carlo samples as Equation (\ref{eq:elbograd}) is intractable. The noise in the ELBO estimate is due to this stochastic approach. Hence, we can consider this and the general stochastic gradient ascent/descent problem as one of estimating a fixed but unknown gradient $\mu = \nabla_\lambda\mathcal{L}$. In consequence, it is feasible to borrow established techniques from statistical estimation theory \cite{Lehmann-1998} for further constraining and regulating the behavior of (\ref{eq:sgd}).

We introduce the following necessary assumption, which is shared with the proof used by \cite{Xu-2019} to demonstrate the variance reduction properties of the reparametrization trick.

\begin{assumption}
    \label{asm:normal}
    Given a sample $z_s$, for $s = 1,2,..., S$, the sample average
    \begin{align*}
        \hat\mu = \frac{1}{S} \sum^{S}_{s = 1}z_s \sim \mathcal{N}(\mu, \sigma^2)
    \end{align*}
    for
    \begin{align*}
    z_s = \nabla_{\lambda} \log q(\theta[s] | \lambda) (\log p(y, \theta[s]) - \log q(\theta[s] | \lambda))
    \end{align*}
    and $S \to +\infty$.
\end{assumption}

Our confidence in the applicability of Assumption \ref{asm:normal} rests in the Central Limit Theorem. We observe that the estimator for the gradient is merely a simple average over independent, identically distributed observations of $z_s$. Thus, for $S \to \infty$ sufficiently large, normality can reasonably be expected to hold.

We see then that BBVI-Naive in Algorithm \ref{alg:bbvi} can be recast as a maximum likelihood estimator, $\hat\mu_{MLE}$, as re-stated in the following theorem.

\begin{theorem}[BBVI as MLE Estimator]
    BBVI-Naive, which we now denote as $\hat\mu_{MLE}$ is the Maximum Likelihood estimator of $\mu = \nabla_\lambda \mathcal{L}$, where   
    \begin{align}
        \hat\mu_{MLE} = \frac{1}{S} \sum^{S}_{s = 1} z_s
    \end{align}
    for $z_s = \nabla_{\lambda} \log q(\theta[s] | \lambda) (\log p(y, \theta[s]) - \log q(\theta[s] | \lambda))$. Furthermore, $\hat\mu_{MLE}$ is unbiased to the true gradient $\mu$.
\end{theorem}

The proof immediately follows from Assumption \ref{asm:normal}. \cite{Ranganath-2014} supports the unbiasedness of this estimator, showing that $E(\hat\mu_{MLE}) = \mu$. However, it is well known \cite{Lehmann-1998, Efron-1973} that for $p > 2$ dimensions, the MLE estimator is dominated in mean square error (MSE) by the James-Stein estimator, specifically the Positive Part James-Stein estimator.

\begin{theorem}[Positive Part James-Stein Estimator]
    \label{thm:jsteinplus}
    The Positive-Part James-Stein estimator $\hat\mu_{JS+}$ given by
    \begin{align*}
        \hat\mu_{JS+} = \bigg(1 - \frac{(p-3) \sigma^2}{|| \bar{z} ||^2} \bigg)^+ \bar{z}
    \end{align*}
    for $\bar{z} = \frac{1}{S} \sum_{s} z_s$ and $(g)^+ = gI_{[0, +\infty)}(g)$ dominates $\hat\mu_{MLE}$ in MSE.
\end{theorem}

The proof for Theorem \ref{thm:jsteinplus} is already a canonical result in estimation theory and can be obtained from \cite{Lehmann-1998}, while an Empirical Bayes approach can be found in \cite{Efron-1973}. The algorithm BBVI-JS+ is summarized in Algorithm \ref{alg:bbvi-js}.

\begin{algorithm}[ht]
    \label{alg:bbvi-js}
    \caption{Positive-Part James-Stein BBVI (BBVI-JS+)}

    \DontPrintSemicolon
    \SetAlgoLined
    \SetKwInOut{Input}{Input}\SetKwInOut{Output}{Output}
    \Input{Model, Monte Carlo Sample Size $S$, convergence threshold $\varepsilon$, learning rate $\rho^t$}
    Initialize $\lambda^0$ randomly, set $t=0$ and $\Delta$ = $\infty$ \;
    \BlankLine
    \While{$\Delta > \varepsilon$}{
      $t = t+1$\;
      // Draw $S$ samples from $q(\theta | \lambda^{t-1})$ \;
      \For{s = 1 to S}{
        $\theta[s] \sim q(\theta | \lambda^{t-1})$ \;
      }
      $\bar{z} = \frac{1}{S} \sum^{S}_{s=1} \nabla_{\lambda} \log q(\theta[s] | \lambda^{t-1}) (\log p(y, \theta[s]) - \log q(\theta[s] | \lambda^{t-1}))$ \;
      $\hat\nabla_\lambda \mathcal{L}(\lambda^{t-1}) = \bigg(1 - \frac{(p-3) \sigma^2}{|| \bar{z} ||^2} \bigg)^{+} \bar{z}$\;
      $\lambda^t = \lambda^{t-1} + \rho^t \hat\nabla_\lambda \mathcal{L}(\lambda^{t-1})$ \;
      $\Delta = \frac{|| \lambda^t - \lambda^{t-1} ||}{|| \lambda^{t-1}||}$ \;
    }
\end{algorithm}

Having defined our proposed estimator, we can now make quantitative statements about its relationship with both the Naive and Rao-Blackwellized versions of BBVI.

\begin{theorem}[Variance Reduction of the James-Stein Estimator]
    \label{thm:varcontrol1}
    Given the Naive BBVI/MLE Estimator $\hat\mu_{MLE}$ and the Positive-Part James-Stein estimator $\hat\mu_{JS+}$, then
    \begin{align*}
        V(\hat\mu_{JS+}) <  V(\hat\mu_{MLE})
    \end{align*}
\end{theorem}

\textit{Proof}. Due to the MSE dominance of $\hat\mu_{JS+}$ over $\hat\mu_{MLE}$, and taking advantage of the Bias-Variance decomposition of MSE, we have that
\begin{align*}
    V(\hat\mu_{JS+}) + \text{Bias}^2(\hat\mu_{JS+}) \leq  V(\hat\mu_{MLE})
\end{align*}
as $\hat\mu_{MLE}$ is an unbiased estimator. From the above proof we find that the James-Stein estimator, when applied to BBVI, should be able to control the sampling distribution of $\hat\nabla\mathcal{L}$ in Equation (\ref{eq:sgd}). The shrinkage factor allows the parameter $\lambda^{t}$ to move when the gradient is relatively small, but forces it to remain near or at its previous value when the gradient explodes.

However, $\hat\mu_{JS+}$ generally performs worse than Rao-Blackwellization $\hat\mu_{RB}$ as shown in the following theorem.

\begin{theorem}
    \label{thm:varcontrol2}
    Given the Positive-Part James-Stein estimator $\hat\mu_{JS+}$ and the Rao-Blackwellized BBVI $\hat\mu_{RB}$
    \begin{align*}
        V(\hat\mu_{RB}) \leq  V(\hat\mu_{JS+})
    \end{align*}
\end{theorem}

\textit{Proof}. Under the factorized case (which is necessary anyway when using $\hat\mu_{RB}$), we can split the summation for some parameter $\lambda_j$
\begin{align*}
\sum^{S}_{s=1} \nabla_{\lambda} \log q(\theta[s] | \lambda^{t-1}) & (\log p(y, \theta[s]) - \log q(\theta[s]|\lambda^{t-1}))  = \\
& \sum^{S}_{s=1} \nabla_{\lambda_j} \log q(\theta_{\lambda_j}[s] | \lambda^{t-1}_{j}) (\log p(y, \theta_{\lambda_j}[s]) - \log q(\theta_{\lambda_j}[s]|\lambda^{t-1}_j)) \\
+ & \sum^{S}_{s=1} \nabla_{\lambda_{-j}} \log q(\theta_{-\lambda_j}[s] | \lambda^{t-1}_{-j}) (\log p(y, \theta_{\lambda_{-j}}[s]) - \log q(\theta_{-\lambda_j}[s]|\lambda^{t-1}_{-j}))
\end{align*}

We can think of the summation in the second term of the right-hand side as an \textit{anti-Markov blanket} to the parameter $\lambda_{j}$. Now, holding the norm $|| \bar{z} ||^2$ constant,
\begin{align*}
    V(\hat\mu_{JS+}) = k \times V(\hat\mu_{RB}) + C
\end{align*}
where
\begin{align*}
    k = \bigg[ \bigg(1 - \frac{(p-3)\sigma^2}{|| \bar{z} ||^2} \bigg)^{+} \bigg]^2
\end{align*}
is a value constrained to $[0,1]$ and
\begin{align*}
    C = k \times V \bigg( \frac{1}{S} \sum^{S}_{s=1} \nabla_{\lambda_{-j}} \log q(\theta_{-\lambda_j}[s] | \lambda^{t-1}_{-j}) (\log p(y, \theta_{\lambda_{-j}}[s]) - \log q(\theta_{-\lambda_j}[s]|\lambda^{t-1}_{-j})) \bigg)
\end{align*}
being the variance of the anti-Markov Blanket of $\lambda_j$ approaches infinity with greater and greater $p$. Hence, the theorem holds for $C \to +\infty$. Our proof has the interesting implication that it is, in fact, possible for $\hat\mu_{JS+}$ to outperform $\hat\mu_{RB}$ in variance, provided that $C \to 0$. In practice this is only possible when there are only very small number of variational parameters, and should be very rare (if it ever occurs) in practice.

Also a consequence of this theorem is an interesting behavior that is achieved when applying the positive-part James-Stein shrinkage factor to the Rao-Blackwellized estimator.

\begin{corollary}
    Suppose we have a Positive-Part Rao-Blackwellized estimator $\hat\mu_{RB+}$ given by
    \begin{align*}
        \hat\mu_{RB+} = \bigg(1 - \frac{(p-3) \sigma^2}{|| \hat\mu_{RB} ||^2} \bigg)^{+} \hat\mu_{RB} 
    \end{align*}
    Then its variance
    \begin{align*}
        V(\hat\mu_{RB+}) \leq  V(\hat\mu_{RB})
    \end{align*}
\end{corollary}

\textit{Proof}. We note that
\begin{align*}
    V(\hat\mu_{RB+}) = k \times V(\hat\mu_{RB})
\end{align*}

This means that a positive-component James Stein estimator applied on the Rao-Blackwellized estimator further constricts the variance of the noisy ELBO gradient estimate.

\subsection{Relationship to Gradient Clipping}
\label{sec:gradclip}

The idea of regulating the path of estimated parameters in stochastic gradient ascent/descent problems (\ref{eq:sgd}) directly through the gradient estimate is not new within the deep learning literature. In the following section, we discuss the connection between the form we have proposed in Algorithm \ref{alg:bbvi-js} with the method of gradient clipping in training deep neural networks.

\begin{figure}[ht]
  \centering
  \includegraphics[width=0.4 \linewidth]{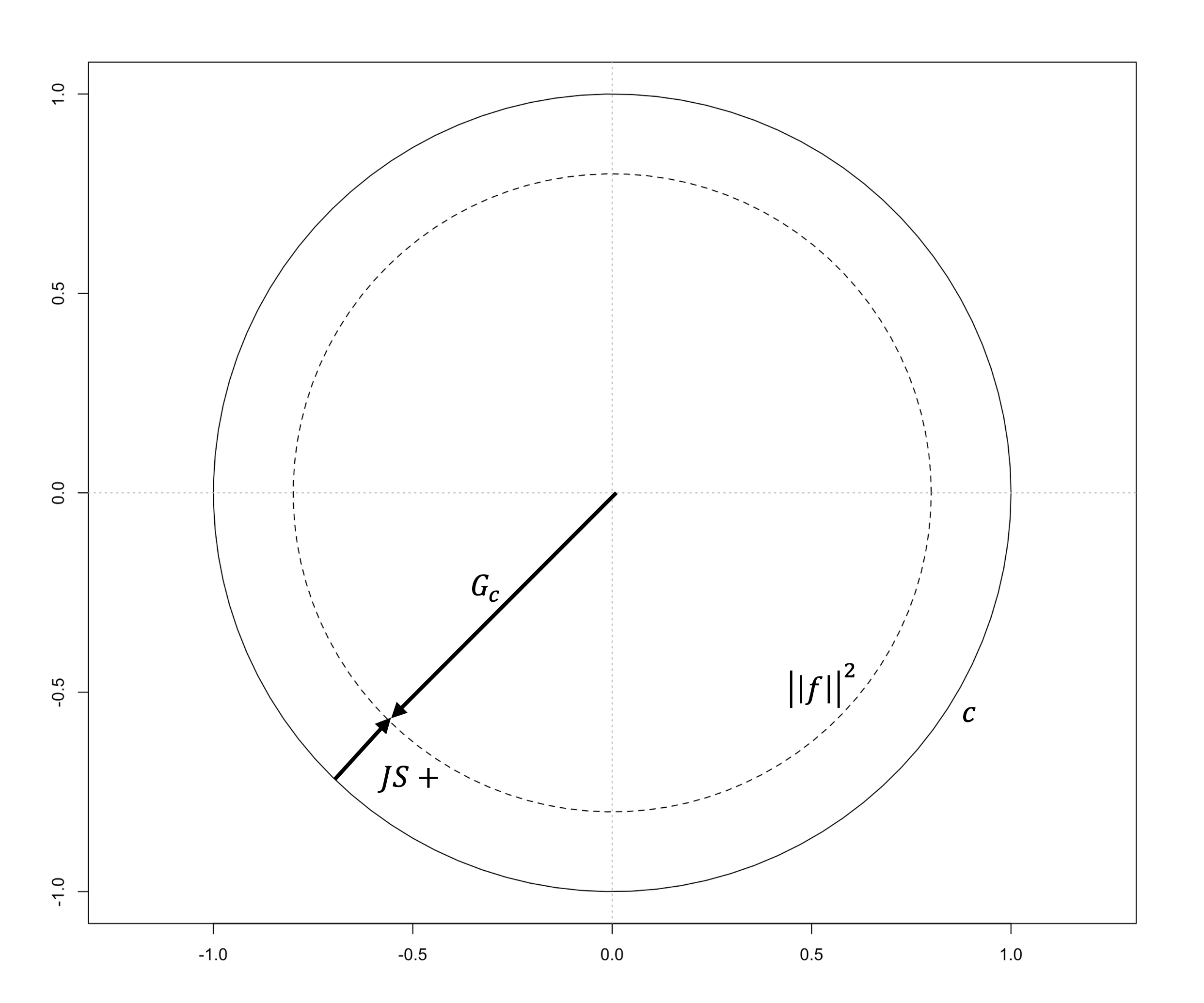}
  \caption{\label{fig:clip} Relationship between gradient clipping and the James-Stein estimator. Gradient clipping $G_c$ preserves values of the gradient only up to $|| f ||^2 \leq c$. The Positive Part James-Stein operator $JS+$ penalizes $||f||^2$ for being close to $c$ and forces it towards zero.}
\end{figure}

Clipping mitigates the issue of exploding gradients, in which estimates of the gradients can become very large during training, leading to instability and straying the parameter updates from convergence. The gradient estimate $\hat\nabla\mathcal{L}$ is effectively constricted by a pre-set radius $c$ (which can be a learnable parameter), such that $\hat\nabla\mathcal{L}$ can only take on values up to $c$. Formally, we can define a clipping function
\begin{align*}
    G_c(f) = \min\bigg(1, \frac{c}{||f||} \bigg)
\end{align*}
and modify Equation (\ref{eq:sgd}) to
\begin{align*}
    \lambda^t = \lambda^{t-1} + \rho^t G_c(\hat\nabla\mathcal{L}(\lambda^{t-1})) \hat\nabla\mathcal{L}(\lambda^{t-1})
\end{align*}

The gradient used in each update step can only be a fraction $\frac{c}{|| \hat\nabla\mathcal{L}(\lambda^{t-1}) ||}$ of the actual gradient value whenever $\hat\nabla\mathcal{L}(\lambda^{t-1}) > c$. In better-behaved iterations when $\hat\nabla\mathcal{L}(\lambda^{t-1}) < c$, the update step is allowed to use the full value of the gradient. With this constraint, clipping prevents the gradients from growing too large, thereby stabilizing the training process. In deep neural networks, clipping limits the influence of any single training sample or layer on the overall parameter updates, leading to more stable training.

We find that it is trivial to suppose a modified form given by
\begin{align}
    \label{eq:clip}
    G_c(f) = \min\bigg(1, \frac{c}{||f||^2} \bigg)
\end{align}
which simply means re-scaling the radius $c$ to be in units of the squared norm of $f$. We now observe the following relationship between clipping and our method.
\begin{theorem}
    The Positive-Part James-Stein estimator $\hat\mu_{JS+}$ represents a reversal of the modified gradient clipping function (\ref{eq:clip}). That is, the shrinkage operator,
    \begin{align*}
        JS(f) = \bigg(1 - \frac{(p-3) \sigma^2}{|| f ||^2} \bigg)^+ = 1 - G_c(f)
    \end{align*}
\end{theorem}

\textit{Proof}. \cite{Efron-1973} show that the shrinkage operator
\begin{align*}
    \bigg(1 - \frac{(p-3) \sigma^2}{|| f ||^2} \bigg)^+ = 1 - \min\bigg(1, 1 - \frac{(p-3) \sigma^2}{|| f ||^2} \bigg)
\end{align*}
which for $c = || f ||^2 - (p-3) \sigma^2$ satisfies the theorem.

This connection is illustrated in Figure \ref{fig:clip}. Gradient clipping is not a shrinkage method, as it does not force the gradient to zero. Instead, clipping is concerned with keeping $f$ within a region such that its squared norm $||f||^2 \leq c$. On the other hand, the James Stein estimator is explicitly a shrinkage method, and it forces the gradient to be as small as possible, imposing a penalty for being close to the limit $c$.

We can then contrast our method from clipping by framing it within a developing framework within the larger field of Bayesian Optimization \cite{Garnett-2023} of keeping updates to stay as close as possible to previous observations \cite{Li-2019}. Whereas clipping is largely a heuristic to prevent exploding gradients, application of the James-Stein estimator represents a prior belief that the correct update step is likely to be small.

\subsection{Further Extensions}
\label{sec:extend}

So far we have maintained ambiguity regarding the learning rate $\rho^t$. In practice, the exact value used for this learning rate contributes significantly to the success of any stochastic gradient ascent/descent problem. It is maintained \cite{Ranganath-2014} that BBVI will converge to its optimal values for a general class of $\rho^t$, in fact requiring only that it follow the Robbins-Monro \cite{Robbins-1951} conditions. In their extensions, it is recommended that $\rho^t$ can follow the AdaGrad algorithm \cite{Duchi-2011, Goodfellow-2016}, as a way of adjusting the learning rate of each parameter dynamically during training. This is achieved by scaling the learning rate for each parameter based on the historical values of the gradient.

However, as a side effect of the biasing done by the James-Stein estimator on the gradient estimates, we find that AdaGrad may be too aggressive in reducing the learning rate towards the latter runs of the algorithm. As the biased gradient means that the algorithm is forced to consider smaller steps, monotonically scaling down the learning rate too fast may prevent the algorithm from reaching an optimal convergence point. Hence, we propose using RMSProp \cite{Goodfellow-2016} instead, which adds an extra parameter $\beta$ that exponentially decreases the impact of the earlier iterations,
\begin{align*}
    G^t & = \beta G^{t-1} + (1 - \beta)g^t(g^t)^T \\
    \rho^t & = \eta \text{diag}(G^t + I \xi)^{-1/2}
\end{align*}
where $g^t$ represents the gradient estimate at the current step $t$, $G^0 = 0$, and $\xi$ is some small positive number to prevent division by zero. The parameters $\beta$ and $\eta$ can both be considered as learnable parameters, though in practice $\beta = 0.9$ or some similarly high number. This decay factor limits the accumulation of historical gradients, and ensures that the learning rates do not become too small over time.

\section{Experiments}
\label{sec:gausmix}

We perform experiments using simulated and benchmark datasets for finite Gaussian mixture models to demonstrate the performance of our proposed algorithm. For demonstrating variance reduction properties, we use simulated data following a univariate mixture of Gaussians and show that the sampling distribution is constricted in BBVI-JS+. We also apply the algorithm to a set of benchmark datasets for clustering tasks provided in the \texttt{FCPS} \cite{FCPS} package. All computations were performed on \texttt{R} version 4.2.3 running under MacOS 14.4.1.

\begin{figure}[ht]
  \centering
  \includegraphics[width= 1 \linewidth]{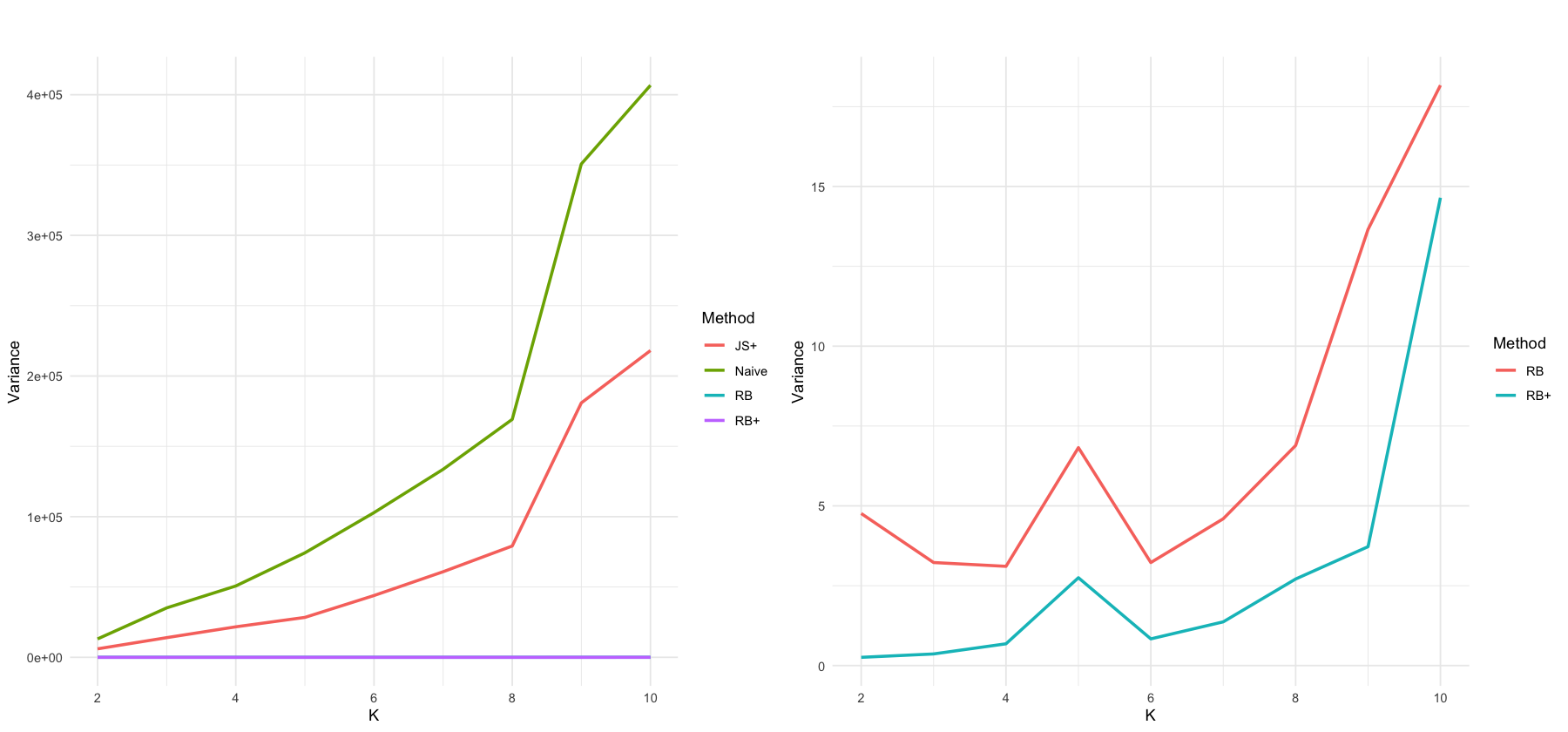}
  \caption{\label{fig:variances} Resulting estimator variances from the Gaussian Mixture experiment with $K = 2$ to $10$ components. BBVI-JS+ produces controlled variances in its sampling distribution for the ELBO gradient compared to BBVI-Naive, but relative to BBVI-RB still grows with the number of parameters. Interestingly, BBVI-RB+ provides even stricter variance control over BBVI-RB.}
\end{figure}

\textbf{Simulated Data.} We follow the formulation presented in \cite{Bishop-2006} except the Dirichlet prior over the mixture components as well as the Inverse-Gamma prior on the variance $\tau^2$. This model has been selected as it permits mimicking the behavior of more complex models easily by adding more components.

Figure \ref{fig:variances} shows the resulting variances of the Naive, James-Stein (JS+), Rao-Blackwellized (RB) as well as James-Stein applied on the Rao-Blackwellized (RB+) BBVI. We find that as the number of components of the mixture distribution, the variance of BBVI-Naive increases linearly as expected. On the other hand, the variance of both RB and RB+ versions remain controlled to several orders of magnitude. Between these two bounds, we have BBVI-JS+ remaining within a level that is found to be between 38 to 46\% of BBVI-Naive in terms of relative efficiency. This confirms our result in Theorems \ref{thm:varcontrol1} and \ref{thm:varcontrol2}. Also in terms of relative efficiency, the performance of BBVI-RB is at around 0.00 to 0.04\% of BBVI-Naive, while BBVI-RB+ is practically at 0.00\%.

\textbf{Benchmarks.} Focusing now on the BBVI-JS+ and BBVI-RB algorithms, we make use of three benchmark datasets found in the \texttt{FCPS} \cite{FCPS} package: EngyTime, Lsun3D, and Tetra. To accommodate these datasets, we extend the univariate gaussian mixture model we used for our simulations to their multivariate counterparts. The results for these benchmarks are provided in Table \ref{tab:benchmarks}.

\begin{table}[ht]
\caption{Time to convergence and fit criteria of Rao-Blackwellized and James-Stein BBVI in three benchmark datasets. For each run, we force the algorithm to take at least 100 iterations before assessing convergence to allow ample warm-up. Due to the differing magnitudes of the gradients, the parameter $\eta$ for RMSProp between the Rao-Blackwellized and James-Stein have been adjusted.}
\label{tab:benchmarks}
\centering
{\small
\begin{tabular}{llrrrrrrrrr}
\hline
\multirow{2}{*}{\textbf{Benchmark}} & \multirow{2}{*}{\textbf{Method}} & \multicolumn{1}{l}{\multirow{2}{*}{\textbf{$\pmb{\eta}$}}} & \multicolumn{1}{l}{\multirow{2}{*}{\textbf{$\pmb{\beta}$}}} & \multicolumn{1}{l}{\multirow{2}{*}{$\mathbf{K}$}} & \multicolumn{1}{l}{\multirow{2}{*}{$\mathbf{p}$}} & \multicolumn{5}{c}{\textbf{Fit}}                                                                                                                                                   \\ \cline{7-11} 
                                    &                                  & \multicolumn{1}{l}{}                              & \multicolumn{1}{l}{}                               & \multicolumn{1}{l}{}                            & \multicolumn{1}{l}{}                            & \multicolumn{1}{l}{\textbf{Iter}} & \multicolumn{1}{l}{\textbf{Time}} & \multicolumn{1}{l}{\textbf{ELBO}} & \multicolumn{1}{l}{\textbf{LogLik}} & \multicolumn{1}{l}{\textbf{DIC}} \\ \hline
\multirow{2}{*}{EngyTime}           & BBVI-RB                          & 1.0                                               & 0.9                                                & 2                                               & 2                                               & 200                               & 13.09                             & -2,268.88                         & -2,516.83                           & 4,535.48                         \\
                                    & BBVI-JS+                           & 0.1                                               & 0.9                                                & 2                                               & 2                                               & 101                               & 4.27                              & -2,231.65                         & -2,363.15                           & 4,459.48                         \\ \hline
\multirow{2}{*}{Lsun3D}             & BBVI-RB                          & 1.0                                               & 0.9                                                & 4                                               & 2                                               & 113                               & 7.81                              & -1,921.70                         & -2,197.60                           & 3,837.53                         \\
                                    & BBVI-JS+                           & 0.1                                               & 0.9                                                & 4                                               & 2                                               & 113                               & 4.69                              & -1,859.93                         & -1,795.25                           & 3,367.49                         \\ \hline
\multirow{2}{*}{Tetra}              & BBVI-RB                          & 1.0                                               & 0.9                                                & 4                                               & 3                                               & 398                               & 24.34                             & -3,197.82                         & -4,238.86                           & 6,389.14                         \\
                                    & BBVI-JS+                           & 0.1                                               & 0.9                                                & 4                                               & 3                                               & 149                               & 5.80                              & -2,470.81                         & -2,812.26                           & 4,556.51                          \\ \hline
\end{tabular}}
\end{table}

Results on the benchmark show that BBVI-JS+ combined with the RMSProp learning rate generally reaches convergence faster than BBVI-RB even with its larger sampling distribution for the ELBO gradient estimate. This performance is attributed to two key factors: because the ELBO gradients are still computed as a whole, there is no need to cycle through the entire dataset for computing the difference factor in Equation (\ref{eq:elbograd}). Moreover, the shrinkage penalty in BBVI-JS+ appear to have allowed the parameter to update slowly and with smaller steps per iteration, keeping the algorithm from performing massive U-Turns in resulting ELBO. Also seen in Table \ref{tab:benchmarks} is that BBVI-JS+ consistently resulted in higher ELBO, and lower DIC than BBVI-RB, although the differences are not very vast. At the very least, these results demonstrate that BBVI-JS+ is able to perform at least at the level of BBVI-RB in coming up with optimal approximations to target posterior densities.

\section{Conclusions}

We have proposed a method of controlling for the variance of the noisy ELBO gradient estimates in Black Box Variational Inference by first casting the stochastic gradient ascent problem as one of estimating a true gradient at each iteration. Borrowing from an established property of multivariate estimators, we proposed a shrinkage operator in the form of the Positive Part James-Stein estimator to bias the gradients towards zero. The result is the inclusion of a prior belief at each iteration that each parameter's update step should be small. Theoretical and empirical results confirm that such a behavior results in narrower sampling distributions for the estimated gradients, and in consequence more stable paths towards optimal values of the variational parameters.

\section*{References}
\bibliographystyle{unsrtnat}

{
    \small
    \bibliography{bbvi}
}

%%%%%%%%%%%%%%%%%%%%%%%%%%%%%%%%%%%%%%%%%%%%%%%%%%%%%%%%%%%%
\clearpage
\appendix

\section{Algorithm Listings}
\label{sec:algorithms}

In order to preserve space, we have left out the complete algorithm listing for two already known variants of BBVI, particularly BBVI-Naive (Algorithm \ref{alg:bbvi}) and BBVI-RB (Algorithm \ref{alg:bbvi-rb}).

\begin{algorithm}[ht]
    \label{alg:bbvi}
    \caption{Naive Black Box Variational Inference (BBVI-Naive) \cite{Ranganath-2014}}

    \DontPrintSemicolon
    \SetAlgoLined
    \SetKwInOut{Input}{Input}\SetKwInOut{Output}{Output}
    \Input{Model, Monte Carlo Sample Size $S$, convergence threshold $\varepsilon$, learning rate $\rho^t$}
    Initialize $\lambda^0$ randomly, set $t=0$ and $\Delta$ = $\infty$ \;
    \BlankLine
    \While{$\Delta > \varepsilon$}{
      $t = t+1$\;
      // Draw $S$ samples from $q(\theta | \lambda^{t-1})$ \;
      \For{s = 1 to S}{
        $\theta[s] \sim q(\theta | \lambda^{t-1})$ \;
      }
      $\hat\nabla_\lambda \mathcal{L}(\lambda^{t-1}) = \frac{1}{S} \sum^{S}_{s=1} \nabla_{\lambda} \log q(\theta[s] | \lambda^{t-1}) (\log p(y, \theta[s]) - \log q(\theta[s] | \lambda^{t-1}))$ \;
      $\lambda^t = \lambda^{t-1} + \rho^t \hat\nabla_\lambda \mathcal{L}(\lambda^{t-1})$ \;
      $\Delta = \frac{|| \lambda^t - \lambda^{t-1} ||}{|| \lambda^{t-1}||}$ \;
    }
\end{algorithm}

\begin{algorithm}[ht]
    \label{alg:bbvi-rb}
    \caption{Rao-Blackwellized BBVI (BBVI-RB) \cite{Ranganath-2014}}

    \DontPrintSemicolon
    \SetAlgoLined
    \SetKwInOut{Input}{Input}\SetKwInOut{Output}{Output}
    \Input{Model, Monte Carlo Sample Size $S$, convergence threshold $\varepsilon$, learning rate $\rho^t$}
    Initialize $\lambda^0$ randomly, set $t=0$ and $\Delta$ = $\infty$ \;
    \BlankLine
    \While{$\Delta > \varepsilon$}{
      $t = t+1$\;
      // Draw $S$ samples from $q(\theta | \lambda^{t-1})$ \;
      \For{$s = 1$ to $S$}{
        $\theta[s] \sim q(\theta | \lambda^{t-1})$ \;
      }
      \For{$j = 1$ to $p$}{
          $\hat\nabla_{\lambda_j} \mathcal{L}(\lambda^{t-1}_{j}) = \frac{1}{S} \sum^{S}_{s=1} \nabla_{\lambda_j} \log q(\theta_{\lambda_j}[s] | \lambda^{t-1}_{j}) (\log p(y, \theta_{\lambda_j}[s]) - \log q(\theta_{\lambda_j}[s]|\lambda^{t-1}_j))$ \;
          $\lambda^t_j = \lambda^{t-1}_j + \rho^t \hat\nabla_{\lambda_j} \mathcal{L}(\lambda^{t-1}_j)$ \;
      }
      $\Delta = \frac{|| \lambda^t - \lambda^{t-1} ||}{|| \lambda^{t-1}||}$ \;
    }
\end{algorithm}

\clearpage

\section{Experimental Setup}
\label{sec:experiments}

All experiments reported in this paper were conducted on an M1 MacBook Air running \texttt{R} version 4.2.3 under MacOS 14.4.1.

\textbf{Simulated Dataset.} The simulated dataset for demonstrating variance reduction in BBVI-JS+ was done by sampling a total of $N = 200$ observations from a finite mixture of Gaussians with $K = 2$ to $10$ components, means and variances deterministically set at $\mu_k = \{-5,-4,-3,-2,-1,0,+1,+2,+3,+4\}$ and common $\sigma^2 = 3$, respectively. During Monte Carlo estimation, a total sample size of $S = 500$ were drawn, and the sampling distributions reported are based on bootstrap samples of size $B = 100$.

Our generative model is given by
\begin{align*}
    \mu_k & \sim \mathcal{N}(0, \tau^2) \\
    z_i  & \sim \text{Categorical}(1/K, ..., 1/K) \\
    y_i | z_{ik} = 1, \mu, \tau^2 & \sim \mathcal{N}(\mu_k, \sigma^2)
\end{align*}

We can propose mean-field variational approximations
\begin{align*}
    q(\mu_k | m_k, s^2_k) & = \mathcal{N}(m_k, s^2_k) \\
    q(z_i | \phi_i) & = \text{Categorical}(\phi_i)
\end{align*}

\begin{figure}[ht]
  \centering
  \begin{multicols}{3}
    \includegraphics[width=\linewidth]{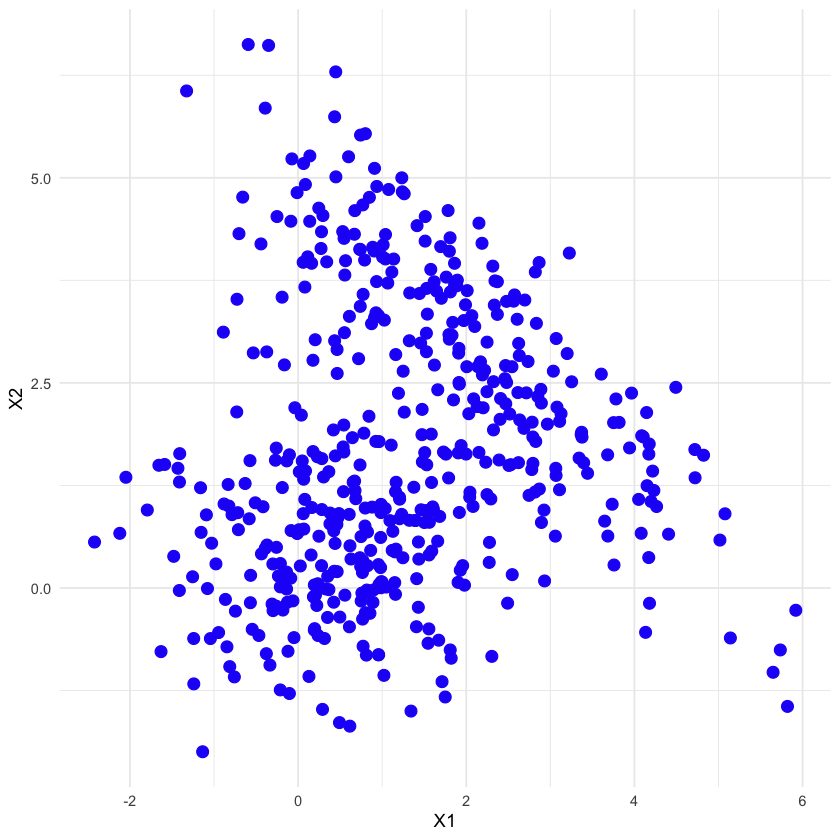}\par 
    \includegraphics[width=\linewidth]{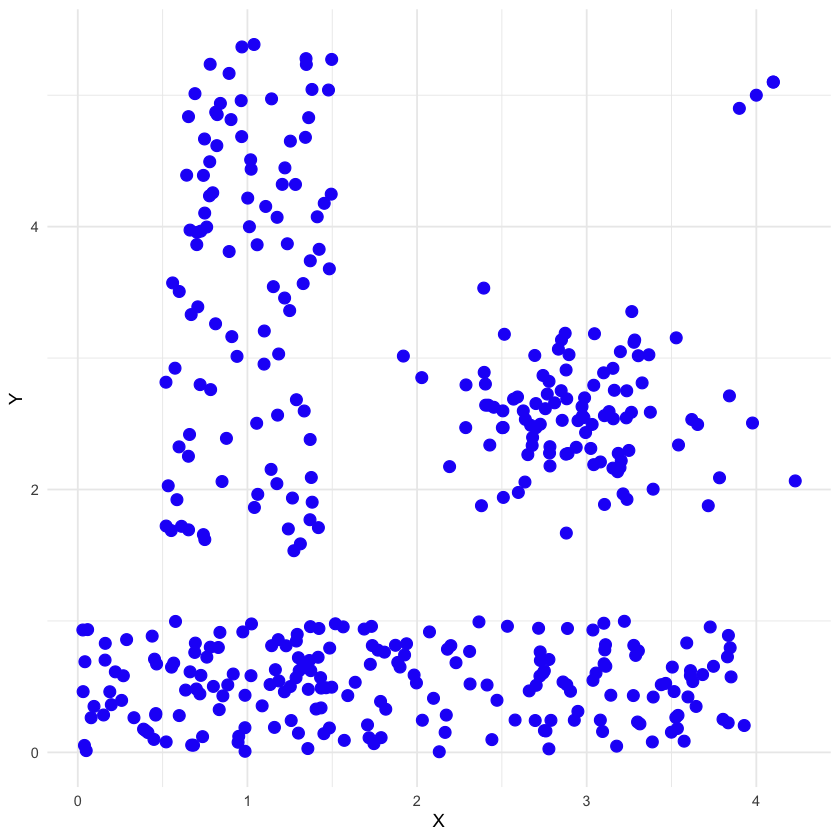}\par 
    \includegraphics[width=\linewidth]{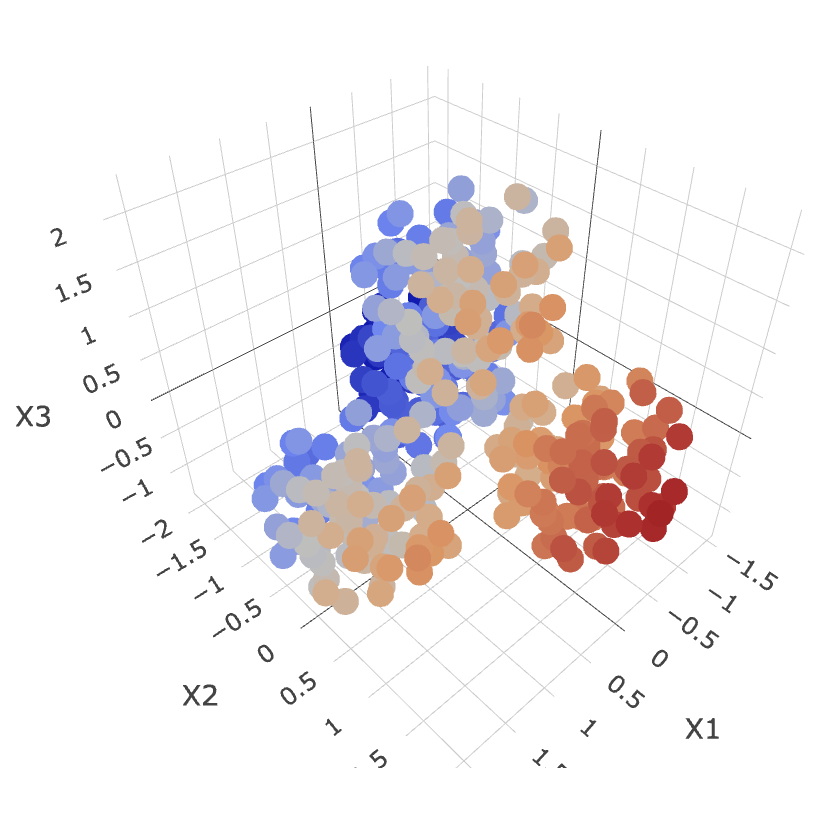}\par 
    \end{multicols}
  \caption{\label{fig:benchmarks} Scatterplots of benchmark datasets from the \texttt{FCPS} package \cite{FCPS}. From left to right: EngyTime, Lsun3D, and Tetra.}
\end{figure}

\textbf{Benchmark Datasets.} To demonstrate convergence and final model fit statistics resulting from using BBVI-JS+, we use three datasets for clustering and Gaussian mixture model tasks provided in the \texttt{FCPS} package \cite{FCPS} for \texttt{R}. Specific information regarding these datasets can be found in the package documentation. A specific note is made for the \texttt{EngyTime} dataset, which originally contains 4,096 observations. To ease computation time, we use a random subset containing 400 randomly selected observations. Scatterplots of the datasets used are in Figure \ref{fig:benchmarks}.

To prevent early convergence of the model at sub-optimal locations, we force the models to run for at least 100 iterations before assessing convergence. This is due to an observed tendency of BBVI in general to first oscillate within sub-optimal solutions, likely as a consequence of initializations, before taking a more consistent trajectory towards maximum ELBO. We note that selection of proper convergence criteria for BBVI is currently an active research problem \cite{Ranganath-2014, Welandawe-2022}. Nevertheless, if a model has already reached convergence before 100 iterations, then it is expected that it should stop within a few iterations after the warm-up. A convergence criterion of $\varepsilon = 0.01$ is used for the \texttt{EngyTime} dataset, and $\varepsilon = 0.1$ for the \texttt{Lsun3D} and \texttt{Tetra} datasets. In each dataset, the convergence criterion used for BBVI-JS+ and BBVI-RB are always the same.

\end{document}